\documentclass[10pt,twocolumn,letterpaper]{article}

\usepackage{cvpr}              % To produce the CAMERA-READY version
\usepackage{graphicx}
\usepackage{float}
 \usepackage{subcaption}
\usepackage{amsmath}
\usepackage{color}
\usepackage{amssymb}
\usepackage[accsupp]{axessibility}
\usepackage{booktabs}
\usepackage{multirow}

\usepackage[pagebackref,breaklinks,colorlinks]{hyperref}

\usepackage[capitalize]{cleveref}
\crefname{section}{Sec.}{Secs.}
\Crefname{section}{Section}{Sections}
\Crefname{table}{Table}{Tables}
\crefname{table}{Tab.}{Tabs.}

\def\cvprPaperID{134} % *** Enter the CVPR Paper ID here
\def\confName{CVPR}
\def\confYear{2023}

\begin{document}

% \title{Super-Resolution Diffusion Models via Implicit Representation}
\title{Implicit Diffusion Models for Continuous Super-Resolution }

\author{
Sicheng Gao{$^{1}$}\thanks{These authors contributed equally.}, Xuhui Liu{$^1$}\footnotemark[1], Bohan Zeng{$^{1}$}\footnotemark[1], Sheng Xu{$^{1}$},
Yanjing Li{$^{1}$}, Xiaoyan Luo{$^{1}$}\\
Jianzhuang Liu{$^2$}, Xiantong Zhen{$^{3}$}, Baochang Zhang{$^{1,4}$}\thanks{Corresponding Author: bczhang@buaa.edu.cn.} \\ 
{$^1$}Beihang University \ {$^2$}Shenzhen Institute of Advanced Technology, Shenzhen, China \\ {$^3$}United Imaging {$^{4}$}Zhongguancun Laboratory, Beijing, China \\
% {\tt\small {bczhang@buaa.edu.cn}}
}

\maketitle
\begin{abstract}
% is a process that attempts to increase the resolution and ultimately the quality of low-resolution images.  It 
% because it stems from the ill-posed many-to-one mapping between high-resolution and low-resolution images.yet it remains challenging 
Image super-resolution (SR) has attracted increasing attention due to its widespread applications.
However, current SR methods generally suffer from over-smoothing and artifacts, and most work only with fixed magnifications.
% Recent generative models can solve the over-smoothing problem caused by  regression-based methods,  but  require large-factor SR model and additional priors. P
% They are still challenged by unnatural artifacts  and fixed resolutions  problems. 
This paper introduces an Implicit Diffusion Model (IDM) for high-fidelity continuous image super-resolution. IDM integrates an implicit neural representation and a denoising diffusion model in a unified end-to-end framework,
where the implicit neural representation is adopted in the decoding process to learn  continuous-resolution representation. 
% We address the fact that direct diffusion processes for SR are confined to fixed super-resolution ratios, by implicitly mapping the data features in the non-discrete domain to learn resolution-continuous representations of images.
% Using a scheduled diffusion denoising procedure, IDM iteratively transforms Gaussian noise into photo-realistic images conditioned on low-resolution input.  
Furthermore, we design a scale-adaptive conditioning mechanism that consists of a low-resolution (LR) conditioning network and a scaling factor. The scaling factor regulates the resolution and accordingly modulates the proportion of the LR information and generated features in the final output, which enables the model to accommodate the continuous-resolution requirement. %Note that 
% Note that IDM does not require extra priors
% By this implicit representation with the conditioning mechanism, IDM can exhibit strong performance easily through a simple iterative denoising process, on various images.
Extensive experiments validate the effectiveness of our IDM and demonstrate its superior performance over prior arts. 
The source code will be available at \url{https://github.com/Ree1s/IDM}.
% Image Super-Resolution,  one of the hottest topics in computer vision,  has achieved a promising performace due to the advancing properties of generative models. However, they suffers from critical over-smoothing, unnatural artifacts and fixed magnification, and most of them still rely on additional priors or complicated two-stage pipelines.  To  exploit the merits of both methods in a unified framework,
% A scaling factor is used for We introduce an implicit image function in the upsampling part of the U-Net architecture, leading to a hierarchical architecture of an alternating denoising model.
% The LR conditioning network efficiently encodes LR images to multi-scale features for the constant denoising steps. And the scaling factor works through Multi-Layer Perceptrons (MLP) to adaptively adjust how much the encoded LR and generated features are expressed.
% There are two types of methods,  regression-based and generative models, each of which has its own advantages and disadvantages. Regression-based methods are intuitive  but  often fail  to achieve high fidelity. generative models can generate realistic perceptual quality but with either redundant training strategies or restricted frameworks.
\end{abstract}

%%%%%%%%% BODY TEXT

\vspace{-2mm}
\section{Introduction}
\label{sec:intro}
% multi-consistent super resolution
Image super-resolution (SR) refers to the task of generating high-resolution (HR) images from given low-resolution (LR) images. It has attracted increasing attention due to its far-reaching applications, such as video restoration, photography, and accelerating data transmission. While significant progress has been achieved recently, existing SR models predominantly suffer from suboptimal quality and the requirement for fixed-resolution outputs, leading to undesirable restrictions in practice.
% While significant progress has been achieved recently with the help of deep neural networks, image SR remains challenging because it is intrinsically an ill-posed problem with many-to-one mapping between  high-resolution and low-resolution images. % \begin{figure}[htpb]
\begin{figure}[t]
%     \centering
%     \includegraphics[width=0.5\textwidth]{cvpr2023-author_kit-v1_1-1/latex/figures/introduction/motivation.pdf}
%     \caption{Qualitative comparison on 8 $\times$ on CelebA-HQ \cite{karras2017progressive}. The results of IDM maintain higher fidelity and more credible identities close to the ground truth, generating more realistic facial components \emph{e.g.}(eyes, teeth, and hair).}
%     \label{fig:face}
% \end{figure}
    \centering
    \includegraphics[width=0.48\textwidth]{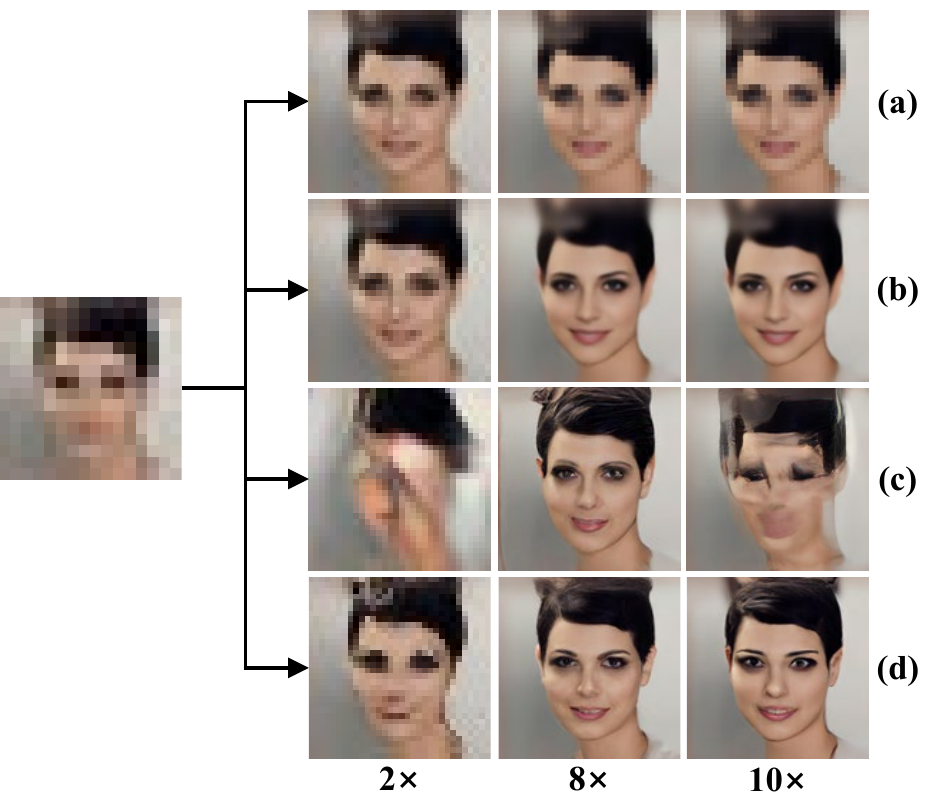}
    \caption{Visual comparison, where training is on 8$\times$ SR and testing on 2$\times$, 8$\times$, and 10$\times$. (a) EDSR \cite{lim2017enhanced} and (b) LIIF \cite{chen2021learning} are regression-based models; (c) SR3 \cite{saharia2022image} and (d) IDM (ours) are generative models. Among them, LIIF and IDM employ the implicit neural representation.}
    % The results of IDM maintain higher fidelity and more credible identities close to the ground truth, generating more realistic facial components \emph{e.g.}(eyes, teeth, and hair).
    \label{fig:intro_implicit}
\end{figure}

\begin{figure*}[t]
      \includegraphics[width=1.\textwidth]{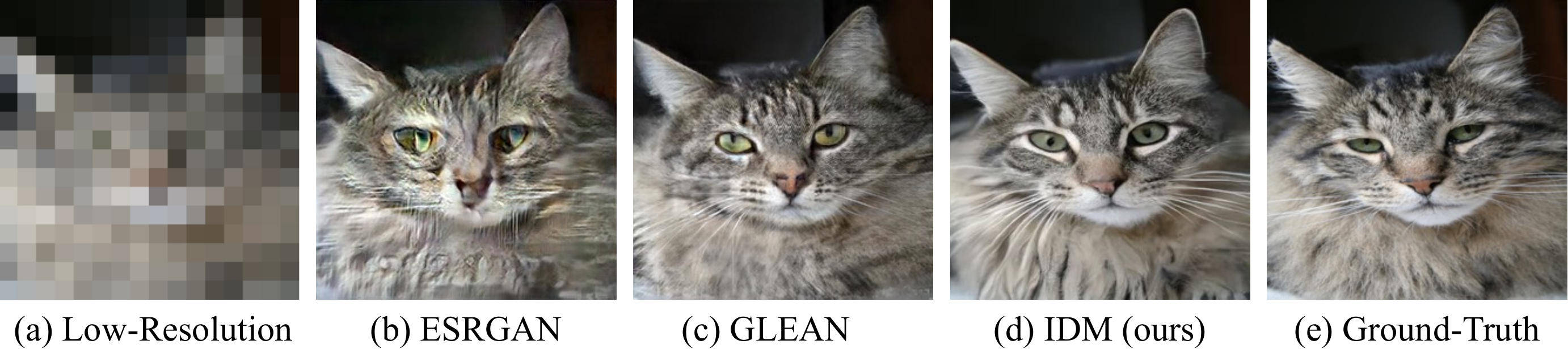}
        \captionof{figure}{\label{intro_pic}
        Examples of 16 $\times$ super-resolution. (a) LR input. (b) ESRGAN \cite{wang2018esrgan} which trains a simple end-to-end structure GAN, and loses the inherent information. (c) GLEAN \cite{chan2021glean} which achieves more realistic details through additional StyleGAN \cite{karras2019style} priors, but still generates unnatural textures and GAN-specific artifacts. (d) With implicit continuous representation based on a scale-adaptive conditioning mechanism, IDM generates the output with high-fidelity details and retains the identity of the ground-truth. (e) The ground-truth.}
\end{figure*}

Regression-based methods \cite{lim2017enhanced,liang2021swinir} offer an intuitive way to establish a mapping from LR to HR images. LIIF \cite{chen2021learning} specifically achieves resolution-continuous outputs through implicit neural representation. However, these methods often fail to generate high-fidelity details needed for high magnifications (see Fig. \ref{fig:intro_implicit}(a) and (b)) since their regression losses tend to calculate the averaged results of possible SR predictions. 
% LIIF \cite{} , but can not address the 
% Current generative SR methods can enrich the detailed texture, while are still subjected to unnatural artifacts and only applicable to fixed magnification ratios.
Deep generative models, including autoregressive \cite{oord2016wavenet,van2016conditional}, GAN-based \cite{karras2017progressive, karras2019style, ledig2017photo, menon2020pulse}, flow-based \cite{dinh2014nice, lugmayr2020srflow} and variational autoencoders (VAEs) \cite{kingma2013auto, vahdat2020nvae}, have emerged as solutions that enrich detailed textures. Still, they often exhibit artifacts and only apply to pre-defined fixed magnifications. 
% ARMs achieve impressive performance, which comes at the price of prohibitive computational requirements, and thus are practically infeasible for HR image generation. Flow-based models and VAEs often generate sub-optimal sample quality than GAN-based models.  
Despite the ability to generate realistic images with high perceptual quality with the help of extra priors, GAN-based models are subject to mode collapse and struggle to capture complex data distributions, yielding unnatural textures. 
%Moreover, most of them still rely on introducing extra GAN priors with elaborately designed modules \cite{} or fine-tuning the priors through complex training strategies \cite{}. 
Recently, Diffusion Probabilistic Models (DMs) \cite{ho2020denoising,sohl2015deep} have been used in image synthesis to improve the fidelity of SR images and have shown impressive performance. Nonetheless, DM-based methods are still limited to fixed magnifications, which would result in corrupted output once the magnification changes (see Fig. \ref{fig:intro_implicit}(c)). Therefore, they turn to a complicated cascaded structure \cite{ho2022cascaded} or two-stage training strategies \cite{rombach2022high, razavi2019generating, guo2022lar} to achieve multiple combined magnifications, or retrain the model for a specific resolution \cite{saharia2022image}, which brings extra training cost. 
% For example, SR3 \cite{saharia2022image} achieve a high-fidelity but only applicable for eightfold magnification, where the generated results do not come up to expectation; CDM \cite{ho2022cascaded} and LDM \cite{rombach2022high} adopt complicated two-stage synthesis pipelines but likewise confined to fixed magnification ratios.

To address these issues, this paper presents a novel Implicit Diffusion Model (IDM) for high-fidelity image SR across a continuous range of resolutions. We take the merit of diffusion models in synthesizing fine image details to improve the
fidelity of SR results and introduce the implicit image function to handle the fixed-resolution limitation.
In particular, we formulate continuous image super-resolution as a denoising diffusion process. We leverage the appealing property of implicit neural representations by encoding an image as a function into a continuous space. When incorporated into the diffusion model, it is  parameterized by a coordinate-based Multi-Layer Perceptron (MLP) to capture the resolution-continuous representations of images better.
% This paper presents a Continuous Image Super-Resolution framework via Implicit Diffusion Refinement (IDM)  for high-fidelity image SR on a continuously arbitrary scale.
% Instead of directly using the diffusion model on SR, we take inspiration from the remarkable properties of implicit neural representation in modeling latent data distributions \cite{} to improve image quality. 
% To this end, we introduce an implicit image function in the upsampling part of the U-Net architecture to learn resolution-continuous representations of images. 

At a high level, IDM iteratively leverages the denoising diffusion model and the implicit image function, which is implemented in the  upsampling layers of the U-Net architecture.  Fig. \ref{fig:intro_implicit}(d) illustrates that IDM achieves continuously modulated results within a wide range of resolutions.
Accordingly, we develop a scale-adaptive conditioning mechanism consisting of an LR conditioning network and a scaling factor. The LR conditioning network can encode LR images without priors and provide multi-resolution features for the iterative denoising steps. The scaling factor is introduced for controlling the output resolution continuously and works through the adaptive MLP to adjust how much the encoded LR and generated features are expressed. 
It is worth noting that, unlike previous methods with two-stage synthesis pipelines \cite{ho2022cascaded,razavi2019generating,esser2021taming} or additional priors \cite{menon2020pulse,chan2021glean, wang2021towards}, IDM enjoys an elegant end-to-end training framework without extra priors.
As shown in Fig. \ref{intro_pic}, we can observe that IDM outperforms other previous works in synthesizing photographic image details.

The main contributions of this paper are summarized as follows:
\begin{itemize}
\item We develop an Implicit Diffusion Model (IDM) for continuous image super-resolution to reconstruct photo-realistic images in an end-to-end manner. Iterative implicit denoising diffusion is performed to learn resolution-continuous representations that enhance the high-fidelity details of SR images. 
% novel arbitrary-scale SR diffusion probabilistic model (IDM) to restore face and natural images continuouslyx` by taking advantage of local implicit neural representation. 
% IDM could naturally connect discrete 2D representation with continuous coordinate information to reconstruct resolution-unconstrained images.
\item We design a scale-adaptive conditioning mechanism to dynamically adjust the ratio of the realistic information from LR features and the generated fine details in the diffusion process.  This is achieved through an adaptive MLP when size-varied SR outputs are needed.
% using realistic identity from LR features, to navigate the generation of fine details in the diffusion process through an adaptive MLP when size-varied SR outputs are needed.
% An effective conditional mechanism based on a controllable arbitrary scaling factor is designed for image-to-image diffusion models. 
% It leverages realistic identity from conditional LR features and generated high fidelity details from the diffusion process through an adaptive MLP when size-varied SR outputs are needed.
\item We conduct extensive experiments on key benchmarks for natural and facial image SR tasks. IDM exhibits state-of-the-art qualitative and quantitative results compared to the previous works and yields high-fidelity resolution-continuous outputs.
% We show our IDM on extensive experiments for both natural and face super resolution tasks. Compared with prior arts, our IDM has achieved extraordinary qualitative and quantitative results and  high-fidelity arbitrary resolution outputs.
\end{itemize}

\begin{figure*}[t]
\centering
\includegraphics[width=0.78\textwidth]{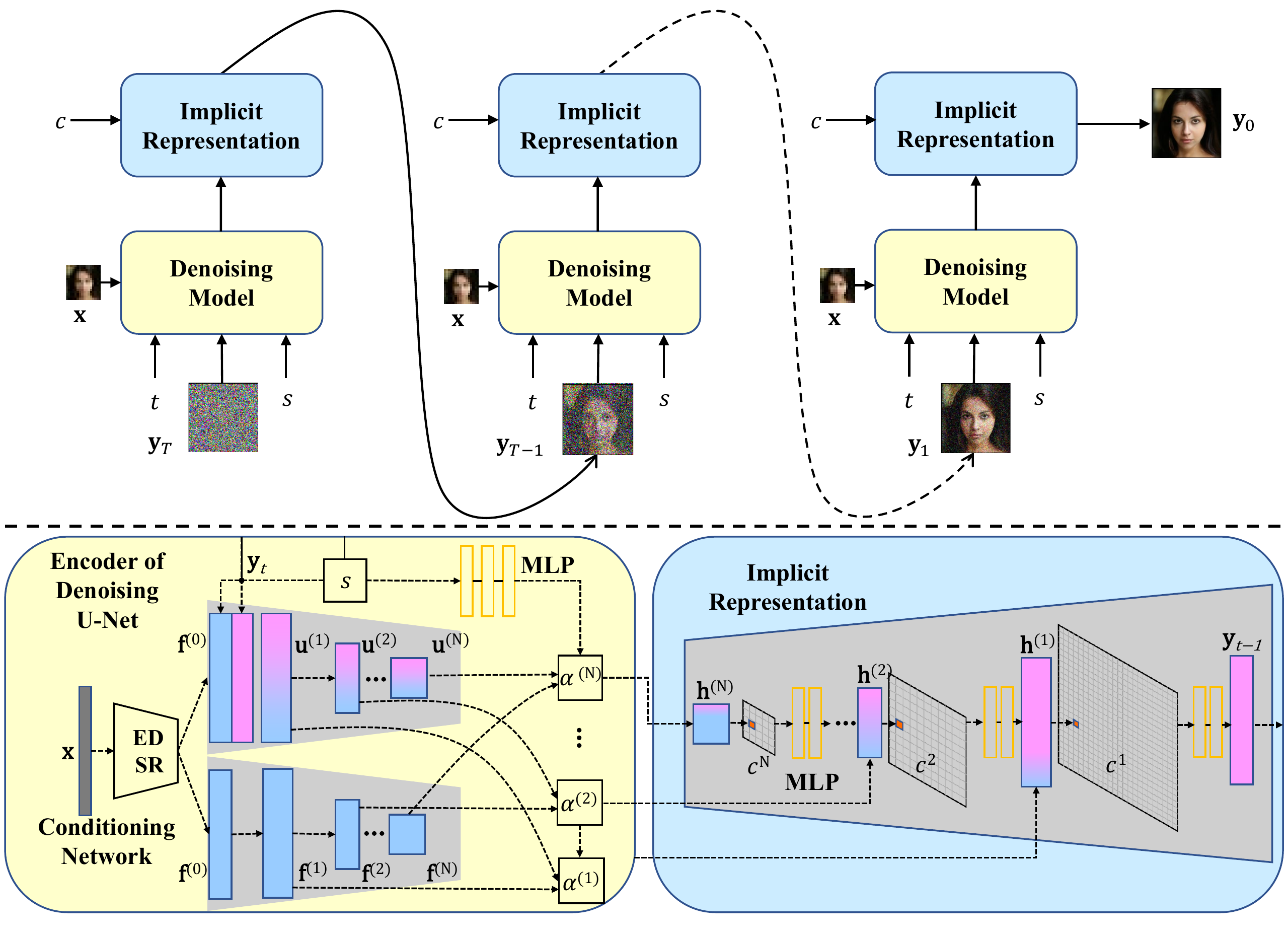} % Reduce the figure size so that it is slightly narrower than the column.
\caption{Overview of the IDM framework. \textbf{Upper Part:} Overall process of the inference. \textbf{Lower Part:} Detailed illustration of a denoising step, where the U-Net decoder is omitted for conciseness. }
\label{fig:overview}
\end{figure*}

\section{Related Work}
\label{sec:Related}
\paragraph{Implicit Neural Representation.} In recent years, implicit neural representations have shown extraordinary capability in modeling 3D object shapes, synthesizing 3D surfaces of the scene, and capturing complicated 3D structures \cite{chabra2020deep, mescheder2019occupancy, mildenhall2021nerf, niemeyer2020differentiable, saito2019pifu, sitzmann2020implicit, sitzmann2019scene}. Particularly, methods based on Neural Radiance Fields (NeRF) \cite{mildenhall2021nerf, barron2021mip} utilize Multi-Layer Perceptrons (MLPs) to render 3D-consistent images with refined texture details. Because of its outstanding performance in 3D tasks, implicit neural representations have been extended to 2D images. Instead of parameterizing 2D shapes with an MLP with ReLU as in early works \cite{rahaman2019spectral,tancik2020fourier}, SIREN \cite{sitzmann2020implicit} employs periodic activation functions to model high-quality image representations with fast convergence. LIIF \cite{chen2021learning} significantly improves the performance of representing natural and complex images with local latent code, which can restore images in an arbitrary resolution. 
However, the high-resolution results generated by LIIF are constrained by prior LR information, resulting in over-smoothing with high-frequency information lost.
% its out-of scale high resolution results have been  replenished from prior LR information. 

In our method, we introduce the denoising diffusion model to yield realistic details missed by LIIF while retaining the superiority of the implicit continuous image function. 
Based on the controllable scaling factor, IDM can dynamically maintain a balance between the LR information and generated fine details while meeting the size-varied requirement of output images.

\paragraph{Generative Image Super-Resolution Models.} 
In image super-resolution, regression-based methods, such as EDSR \cite{lim2017enhanced}, RRDB \cite{wang2018esrgan}, and SWinIR \cite{liang2021swinir}, directly learn a mapping from LR to HR images with an MSE loss. Based on these algorithms, \cite{hu2019meta,chen2021learning,lee2022local} further achieve continuous image super-resolution with meta-learning or implicit neural representation. While impressive PSNR results have been shown, they often suffer from duller edges and over-smoothing details in perceptual outputs. 
On the other hand, GAN-based and flow-based models, variational autoencoders (VAEs), and autoregressive models (ARMs) have been  proposed to improve the fine details of SR images. SRGAN \cite{ledig2017photo} uses an adversarial loss and the perceptual loss \cite{zhang2018unreasonable}, rather than a pixel-wise loss (\emph{e.g.}, L2 loss), to optimize the output. SFTGAN \cite{zhang2019sftgan} and GLEAN \cite{chan2021glean} design new structures to fuse semantics and StyleGAN \cite{karras2019style} priors to generate rich and realistic texture features. Moreover, flow-based models \cite{lugmayr2020srflow, liang2021hierarchical} and VAEs \cite{kingma2013auto, vahdat2020nvae} introduce normalization flow and stochastic variational inference into image generation, respectively, but their sample quality underperforms GAN-based methods. Despite the strong performance in learning complex distributions, ARMs \cite{oord2016wavenet,van2016conditional} are limited to low-resolution images because of the high training cost and sophisticated sequential sampling process.
% Because of the mismatched training cost\textbf{} on pixel-level image representations, applications of ARMs \cite{} are limited in significant measure regardless of rich distributions.

Recently, Diffusion Probabilistic Models (DMs) \cite{ho2020denoising} have shown state-of-the-art results in image and speech synthesis \cite{chen2021wavegrad}, and time series forecasting \cite{pmlr-v139-rasul21a}, for example. Likewise, some diffusion frameworks have been applied to low-level vision tasks. For example, SR3 exhibits impressive performance on image SR after repeated refinement, LDM \cite{rombach2022high} employs the cross-attention conditioning mechanism for generating high-resolution images, and CDM \cite{ho2022cascaded} introduces the class condition and a cascaded structure to achieve realistic multi-resolution results. However, some drawbacks of existing models remain to be solved, including but not limited to unnatural artifacts, fixed magnification ratios, and complicated two-stage pipelines. 
In this paper, IDM combines the merits of diffusion models and implicit neural representations in a practical end-to-end framework, thereby obtaining photo-realistic SR images with continuous resolutions. 

% In view of high fidelity outputs and uncomplicated training tricks of DMs which outperform other generative models, we apply a diffusion framework in the image super-resolution field. To alleviate the resolution-fixed problems of above methods, our IDM focuses on generating high-quality SR images with implicit 2D image representations and is unique by the adaptive ratio between LR features and generative textures. Avoiding previous two-stage image synthesis pipelines \cite{ho2022cascaded,vqvae,vqgan} and additional priors \cite{pulse,chan2021glean, gfpgan}, our IDM adopts an end-to-end framework with no additional priors and achieves more realistic results on both face images and natural scenes.

\section{Method}
\label{sec:method}
% Prevailing GAN inversion methods \cite{menon2020pulse, gu2020image, pan2021exploiting} either present sophisticated CNN modules to modify fixed GAN priors or design specific training strategies to finetune the priors and are prone to unnatural artifacts and loss of details. 
% Recently, DMs have shown impressive performance on high-fidelity generation, but they still adopt inefficient two-stage pipelines and are limited to fixed magnification ratios.
% through cascaded pipelines or fixed structures
% employ complicated CNN modules to use fixed GAN prior or finetune the GAN prior with an additional . Recently, diffusion models have focused on high-fidelity samples through cascaded pipelines or fixed structures, nonetheless, this simple technique requires expensive computing resources.
This section presents the IDM approach, a simple end-to-end framework with an effective scale-adaptive conditioning mechanism and an implicit diffusion process, to generate high-fidelity resolution-continuous outputs. The architecture of IDM is shown in Fig \ref{fig:overview}. 

\subsection{Problem Statement}
\label{sec:overview}
Given an LR-HR image pair denoted as $(\mathbf{x}_i, \mathbf{y}_i)$ and a scaling factor $s$, where $\mathbf{x}_i$ is degraded from $\mathbf{y}_i$ and $s$ controls the resolution of the output in a continuous manner, IDM aims to learn a parametric approximation to the data distribution $p(\mathbf{y} \mid \mathbf{x})$ through a fixed Markov chain of length $T$. Following \cite{saharia2022image}, we define the forward Markovian diffusion process $q$ by adding Gaussian noise as:
\begin{equation}
\begin{aligned}
q\left(\mathbf{y}_{1: T} \mid \mathbf{y}_0\right) &=\prod_{t=1}^T q\left(\mathbf{y}_t \mid \mathbf{y}_{t-1}\right),  \\
q\left(\mathbf{y}_t \mid \mathbf{y}_{t-1}\right) &=\mathcal{N}\left(\mathbf{y}_t \mid \sqrt{1-\beta_t} \mathbf{y}_{t-1},\beta_t \mathbf{I}\right),
\end{aligned}
\end{equation}
where $\beta_t \in(0,1)$ are the variances of the Gaussian noise in $T$ iterations. Given $\mathbf{y}_{0}$, the distribution of $\mathbf{y}_{t}$ can be represented by: 
\begin{equation}
\begin{aligned}
q\left(\mathbf{y}_t \mid \mathbf{y}_{0}\right) &=\mathcal{N}\left(\mathbf{y}_t \mid \sqrt{\gamma_t} \mathbf{y}_{0}, (1-\gamma_t) \mathbf{I}\right),
\end{aligned}
\end{equation}
where $\gamma_t=\prod_{i=1}^t \left( 1 - \beta_i\right)$.
% Meanwhile, the arbitrary scaling factor $s$ in the range of super-resolution training tasks decides the continuous resolution of denoising results cooperated with our coordinate-based MLP implicit function. On the other hand, $s$ is used to control the ratio between LR realistic features and generated missing details in the diffusion process of IDM. 
In the inverse diffusion process, IDM learns the conditional distributions $p_\theta\left(\mathbf{y}_{t-1} \mid \mathbf{y}_t, \mathbf{x}\right)$ to denoise the latent features sequentially during training. Formally, the inference process can be conducted as a reverse Markovian process from Gaussian noise $\mathbf{y}_T \sim \mathcal{N}(\mathbf{0}, \mathbf{I})$ to a target image $\mathbf{y}_0$ as:
% Based on learned conditional distributions $p_\theta\left(\mathbf{y}_{t-1} \mid \mathbf{y}_t, \mathbf{x}\right)$, our IDM inferences from a pure noise image $\mathbf{y}_T \sim \mathcal{N}(\mathbf{0}, \mathbf{I})$ to a target image $\mathbf{y}_0$ through the reverse Markovian process:
\begin{equation}
\begin{aligned}
p_\theta\left(\mathbf{y}_{0: T} \mid \mathbf{x}\right) &=p\left(\mathbf{y}_T\right) \prod_{t=1}^T p_\theta\left(\mathbf{y}_{t-1} \mid \mathbf{y}_t, \mathbf{x}\right), \\
p\left(\mathbf{y}_T\right) &=\mathcal{N}\left(\mathbf{y}_T \mid \mathbf{0}, \mathbf{I}\right), \\
p_\theta\left(\mathbf{y}_{t-1} \mid \mathbf{y}_t, \mathbf{x}\right) &=\mathcal{N}\left(\mathbf{y}_{t-1} \mid \mu_\theta\left(\mathbf{x}, \mathbf{y}_t, t\right), \sigma_t^2 \mathbf{I}\right).
\end{aligned}
\end{equation}
As shown in Fig. \ref{fig:overview}, we adopt a U-Net architecture as the denoising model similar to the vanilla DDPM \cite{ho2020denoising} that encodes the noisy image $\mathbf{y}_t$ into multi-resolution feature maps $\mathbf{u}^{\left(i\right)}$, where $i \in\{1, \cdots, N\}$, and $N$ is the number of depths in the U-Net backbone. Meanwhile, we introduce the implicit image function in the decoding part of the U-Net to generate realistic resolution-continuous images. IDM unifies the iterative diffusion refinement process and the implicit image function in an end-to-end framework.  
% As shown in  
%  To complete the reverse process, our denoising network is similar to U-net backbone \cite{} of vanilla DDPM \cite{} that encodes the noise image $\mathbf{y}_t$ into multi-resolution feature maps $\mathbf{u}^i$.
\subsection{Scale-Adaptive Conditioning Mechanism}
\label{condition}
\paragraph {LR Conditioning Network.}
\label{sec:encoder}
Inspired by GLEAN \cite{chan2021glean} and GCFSR \cite{he2022gcfsr}, we utilize a CNN, which is stacked by convolutional layers with a bilinear filtering downsampling operation and a leaky ReLU \cite{maas2013rectifier} activation, as the conditioning network to extract conditioning features in multiple resolutions from the LR image.
% is designed as a CNN with a stride of 2 to extract features from the LR image.
To accomplish this, we first employ EDSR \cite{lim2017enhanced} to establish the initial LR feature $\mathbf{f}^{\left(0\right)}$ and make its resolution the same as $\mathbf{y}_t$'s by bilinear interpolation. Then, we concatenate $\mathbf{f}^{\left(0\right)}$ and $\mathbf{y}_t$ and feed the result into the U-Net for preliminary conditional guidance. Meanwhile, $\mathbf{f}^{\left(0\right)}$ is also sent to the CNN, where the feature is progressively downsampled as:
% The initial feature from is denoted by . Then, we obtain features whose resolution is progressively reduced as:, \quad i \in\{1, \cdots, N\}
\begin{equation}
\mathbf{f}^{\left(i\right)}=\text{Conv}\left(\mathbf{f}^{\left(i-1\right)}\right),
\end{equation}
where \text{Conv} denotes the convolution layer with the bilinear filtering downsampling operation and a leaky ReLU activation. Unlike GAN-based methods that rely on additional priors \cite{chan2021glean,wang2021towards}, our conditioning network only provides encoded multi-resolution features. It sends them into the U-Net without extra priors to model latent representations.
% This is because we aim to substitute style modulations \cite{styleGAN2} by adding encoded features into the denoising function. 

\paragraph{Scaling Factor Modulation.} 
To lift the limitation of fixed magnification ratios, we introduce a scaling factor $s$ as a condition for the diffusion process to enable magnification with continuous resolution. We first define an interval $(1, M]$, where $M$ is the maximum magnification ratio, and randomly select  $s$ from the interval during training. We then reshape $\mathbf{y}_{t}$ according to $s$ to control the resolution of generated images, as shown in the yellow part of Fig. \ref{fig:overview}.
The scaling factor $s$ is used to adjust the ratio of the original input information $\mathbf{f}^{i}$ from the conditioning network and the output $\mathbf{u}^{i}$ from the denoising network.
% To fit in the resolution-continuous diffusion process,  is designed to adjust the original input information $f^{i}$ from the conditioning network and the creating output texture $u^{i}$ from the denoising network. 
As shown at the bottom  of Fig. \ref{fig:overview}, unlike the cross-attention mechanism \cite{rombach2022high} and the concatenating operation \cite{li2022srdiff}, we map $s$ to a set of scaling vectors $\mathbf{\alpha}=\left\{\alpha_{1}^{(1)},\alpha_{2}^{(1)},\ldots,\alpha_{1}^{(i)},\alpha_{2}^{(i)},\ldots, \alpha_{1}^{(N)},\alpha_{2}^{(N)}\right\}$ with an adaptive MLP, where $i$ represents the depth index with different resolution outputs from the conditioning network and the denoising network.  Next $\alpha_{1}^{(i)}$ and $\alpha_{2}^{(i)}$ are normalized by the L2 norm, and then used to modulate  $\mathbf{f}^{\left(i\right)}$ and  $\mathbf{u}^{\left(i\right)}$ channel-wisely and fuse them adaptively. In general, we conduct the modulation process with the scaling factor $s$ as follows:
% $\alpha_{1}^{(i)},\alpha_{1}^{(i)} \in {\mathbb{R}^{c(i)}}$ and $c(i)$ is the dimension of feature channel.
\begin{center}
\vspace{-5mm}
\begin{align}
    {\mathbf{\alpha }} = Reshape(\operatorname{MLP} (s)), \\
    {\bar{\alpha}_{1}^{(i)}} = \frac{\left|\alpha_{1}^{(i)}\right|}{\sqrt{\alpha_1^{(i)^2}+\alpha_2^{(i)^2}+\delta}}, \\
    {\bar{\alpha}_{2}^{(i)}} = \frac{\left|\alpha_{2}^{(i)}\right|}{\sqrt{\alpha_1^{(i)^2}+\alpha_2^{(i)^2}+\delta}},
\end{align}
\end{center}
\begin{equation}
\begin{aligned}
    {\mathbf{h}^{(i)}} =\bar{\alpha}_{\text {1}}^{(i)} \cdot \mathbf{f}^{(i)}+\bar{\alpha}_{\text {2}}^{(i)} \cdot \text{Concat}\left({\mathbf{u}_{\text{up}}}^{(i)},{\mathbf{u}_{\text{down}}}^{(i)}\right),
\end{aligned}
\end{equation}
where $\delta=1 e-8$ to avoid zero denominators, and ${\mathbf{u}_{\text{up}}}^{(i)}$ and ${\mathbf{u}_{\text{down}}}^{(i)}$ are the feature maps from the decoder and encoder of the U-Net, respectively. The modulation result $\mathbf{h}^{(i)}$ is shown in Fig. \ref{fig:overview}.
% and ${u_{\text{up}}}^{(i)}$, ${u_{\text{down}}}^{(i)}$ denote encoded and decoded feature maps from the U-net backbone.
% , \quad \in\{1, \cdots, N\}

\begin{table}[]
\caption{Datasets used in our experiments.}
\begin{tabular}{c|c|c}
\label{tab:dataset}
                     & Training      & Testing           \\ \hline
% Human faces(Bicubic) & FFHQ \cite{karras2019style}      & CelebA-HQ \cite{karras2017progressive}     \\
Human faces & FFHQ \cite{karras2019style}      & CelebA-HQ \cite{karras2017progressive}     \\
General scenes       & DIV2K \cite{agustsson2017ntire}     & DIV2K-validate \cite{agustsson2017ntire} \\
Cats                 & LSUN-train \cite{yu2015lsun} & CAT \cite{zhang2008cat}           \\
% Cars                 & LSUN-train & Cars           \\
Bedrooms             & LSUN-train \cite{yu2015lsun} & LSUN-validate \cite{yu2015lsun} \\
Towers               & LSUN-train \cite{yu2015lsun} & LSUN-validate \cite{yu2015lsun} \\
% Church               & LSUN-train \cite{yu2015lsun} & LSUN-validate \cite{yu2015lsun}
\end{tabular}
\end{table}

\subsection{Implicit Neural Representation}
\label{sec:ddpm}
% \paragraph{Implicit Continuous Neural Representations.} 
Considering that prevailing SR methods are often burdened by a complicated cascaded pipeline \cite{ho2022cascaded} or two-stage training strategies \cite{rombach2022high, razavi2019generating} to produce outputs with multiple resolutions, we innovate the implicit neural representation to learn continuous image representations,
simplifying IDM.
As shown in the blue box in Fig. \ref{fig:overview}, we insert several coordinate-based MLPs into the upsampling of the U-Net architecture to parameterize the implicit neural representations, which can restore LR images with high-fidelity quality in a continuous scale range.
% To alleviate cascaded pipelines \cite{} or two-stage complex training strategies \cite{}.
% With the help of the scaling factor $s$, we can obtain continuous coordinates of multi-resolution features from the denoising network. 
Like LIIF \cite{chen2021learning}, with the assumed continuous coordinates of multi-resolution features $c=\left\{c^{\left(1\right)}, \ldots, c^{\left(i\right)}, \ldots, c^{\left(N\right)}\right\}$ as a reference, which is obtained from the denoising network using the scaling factor $s$, we input the current features around the coordinates and then calculate the target features. Given the features $\mathbf{h}^{\left(i+1\right)}$ and the corresponding coordinates $c^{\left(i+1\right)}$, we formulate the implicit representation process as follows:
\begin{equation}
\mathbf{u}_{\text{up}}^{\left(i\right)}=D_i\left(\hat{\mathbf{h}}^{\left(i+1\right)}, c^{\left(i\right)}-\hat{c}^{\left(i+1\right)}\right),
\end{equation}
where $D_i$ is a 2-layer MLP with hidden dimensionality 256, and $\hat{\mathbf{h}}^{\left(i+1\right)}$ and $\hat{c}^{\left(i+1\right)}$ are interpolated by calculating the nearest Euclidean distance from $\mathbf{h}^{\left(i+1\right)}$ and $c^{\left(i+1\right)}$ in the $(i+1)$-th depth, respectively. 
% Note that the corresponding coordinates  are gradually expanded in the decoder of the U-Net.
% , \quad i \in\{1, \cdots, N\}

\subsection{Optimization} 
% The denoising Model is employed to infer the target image $\mathbf{y}_0$. 
IDM aims to infer the target image $\mathbf{y}_0$ with a sequence of denoising steps.
To this end, we optimize the denoising model $\epsilon_\theta$ which is equivalent to restoring the target image $\mathbf{y}_0$ from a noisy target image $\tilde{\mathbf{y}}_t=\sqrt{\gamma_t} \mathbf{y}_0+\sqrt{1-\gamma_t} \epsilon$. Meanwhile, to achieve resolution-continuous outputs, the denoising model $\epsilon_\theta\left(\mathbf{x}, t,s, \tilde{\mathbf{y}}_t, \gamma_t\right)$ should apply to arbitrary scales through training while ensuring the validity of the predicted noise $\epsilon$. To conclude, we optimize the denoising network with
\begin{equation}
\mathbb{E}_{(\mathbf{x}, \mathbf{y} )} \mathbb{E}_{\epsilon, \gamma_t, t,s}\left\|\epsilon- \epsilon_\theta(\mathbf{x},t, s, \tilde{\mathbf{y}}_t, \gamma_t)\right\|_1^1,
\end{equation}
where $ \epsilon \sim \mathcal{N}(\mathbf{0}, \mathbf{I})$, $t \sim \{1, \cdots, T\}$, $s \sim \mathcal{U}(1,M]$, and $(\mathbf{x}, \mathbf{y})$ is sampled from the training set of LR-HR image pairs.
% \begin{equation}
% \mathbb{E}_{(\mathbf{x}, \mathbf{y} )} \mathbb{E}_{\epsilon, \gamma, s \sim  $ s \sim \mathcal{U}(0,16)$ \mathcal{U}(0,16)}\left\|\epsilon- \epsilon_\theta(\mathbf{x}, s, \tilde{y}, \gamma)\right\|_1^1
% \end{equation}

\begin{figure}[htpb]
    \centering
    \includegraphics[width=0.5\textwidth]{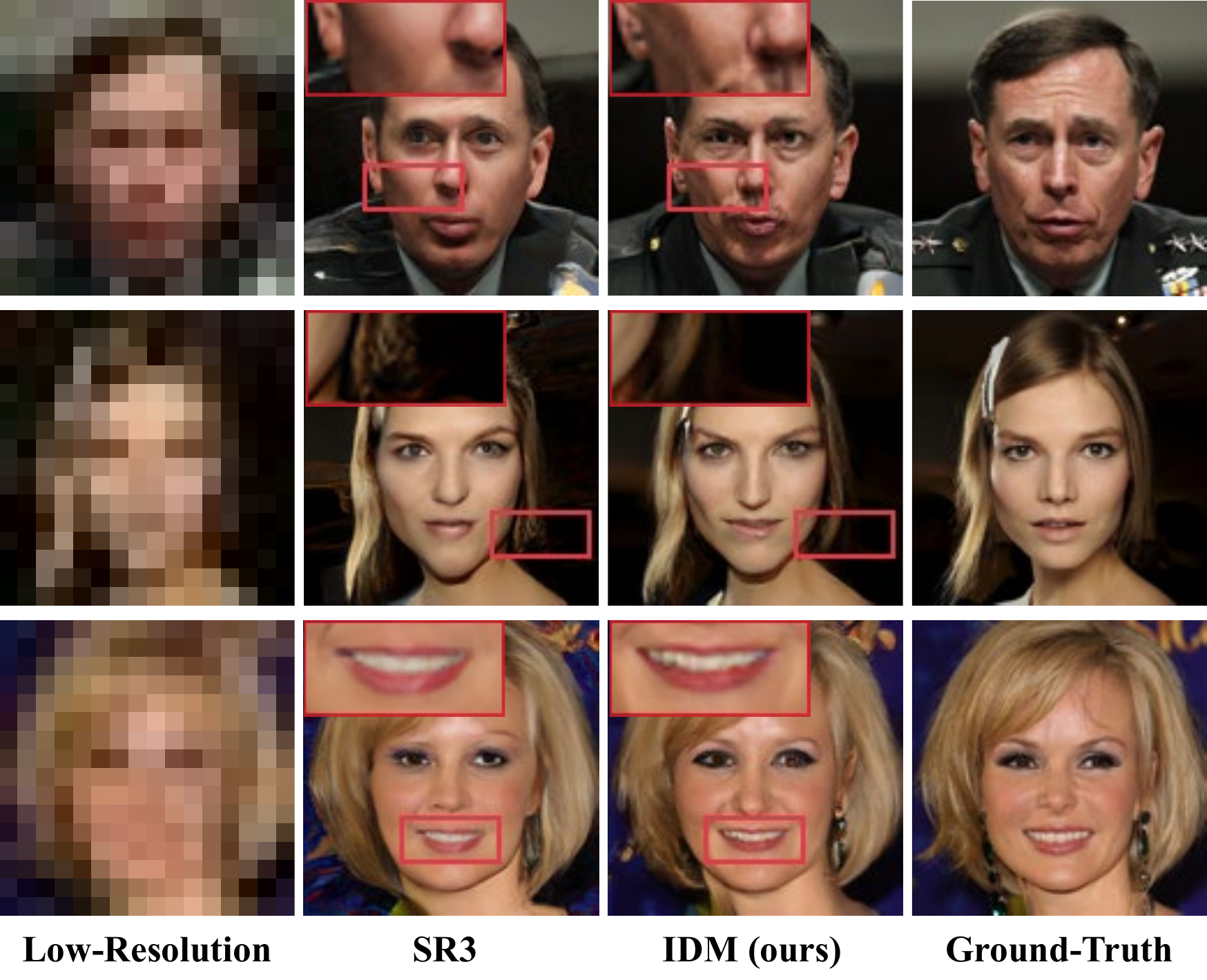}
    \caption{Qualitative comparison on 8$\times$ SR on CelebA-HQ \cite{karras2017progressive}. The results of IDM maintain higher fidelity and more credible identities close to the ground-truth, generating more realistic facial components (\emph{e.g.,} eyes, teeth, and hair).}
    \label{fig:face}
    \vspace{-3mm}
\end{figure}

\section{Experiments}
In addition to the extensive experiments described in this section, we also provide more results with more magnifications and resolutions in the supplementary materials.
\label{sec:experiments}
\subsection{Implementation Details} 
\paragraph{Datasets.}
We conduct our experiments on face datasets, natural image datasets, and a general scene dataset (DIV2K \cite{agustsson2017ntire}), which are listed in Table \ref{tab:dataset}. For face datasets, we train and evaluate IDM on FFHQ \cite{karras2019style} and CelebA-HQ \cite{karras2017progressive}, respectively, which is the same as with SR3 \cite{saharia2022image}. We use DIV2K for general scenes to compare with various state-of-the-art (SOTA) methods based on other generative models. Finally, similar to GLEAN \cite{chan2021glean}, we train and test our model on the LSUN \cite{yu2015lsun} dataset.
\paragraph{Training Details.}
We train our IDM in an end-to-end manner. We set a milestone to 1M iterations, where the training is with a fixed downsampling scale to $M$$\times$, and after the milestone, the training is conducted for 0.5 M iterations with HR images randomly resized according to the uniform distribution $\mathcal{U}(1,M)$.
% and the later conduct 0.5M training steps of randomly resizing the HR images with a uniform distribution $\mathcal{U}(1,M)$.
% two stages. In the first stage, we fix the down-sampling space scale to $\times 16$ and train for 1M steps. In the second stage, we randomly resize the HR images with a uniform distribution $\mathcal{U}(1,16)$  over 0.5M training steps. 
Following the vanilla DDPM \cite{ho2020denoising}, we use the Adam optimizer with a fixed learning rate of 1e-4 for the former and 2e-5 for the latter. We utilize a dropout rate of 0.2 and two 24GB NVIDIA RTX A5000 GPUs for all experiments.
% We  for all . 

\subsection{Qualitative Comparisons}
\begin{figure}[t]
\centering
\includegraphics[width=0.5\textwidth]{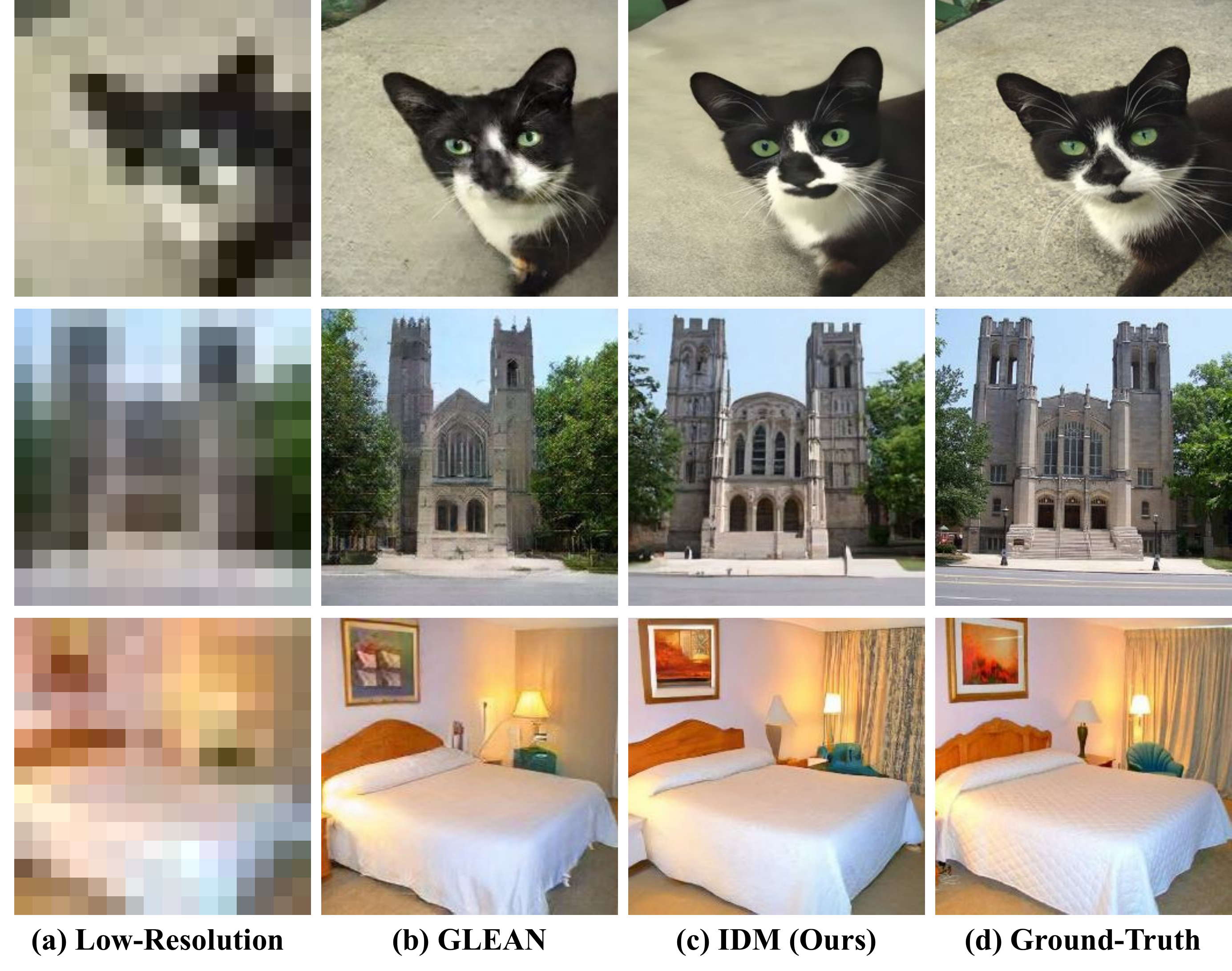} % Reduce the figure size so that it is slightly narrower than the column.
\caption{Results of 16$\times$ SR on the LSUN dataset. IDM achieves more consistent textures with the ground-truth.}
\label{fig:lsun}
% \vspace{-3mm}
\end{figure}

We conduct qualitative comparisons with SOTA methods on both face and natural image SR.
\vspace{-1mm}
\paragraph{Face Super-Resolution.} Fig. \ref{fig:face} shows the qualitative comparison with the SOTA DM-based method (SR3) on the  16$\times$16 $\rightarrow$ 128$\times$128 face SR task. Although both SR3 and IDM can improve the diversity of generated outputs, SR3 loses many face attributes, so it is quite different between the identities of the SR3 outputs and the ground-truth. For instance, the teeth and eyes are discrepant, and the wrinkles are not retained. In contrast, IDM maintains the identities and high-fidelity face details.
\vspace{-1mm}
\paragraph{Natural Image Super-Resolution.} Fig. \ref{fig:lsun} shows the qualitative comparison with the SOTA GAN-based method (GLEAN) on the 16$\times$ natural image SR task, including Cats, Towers, and Bedrooms. We directly take the examples provided in GLEAN for comparison. Although GLEAN can produce realistic SR results, it does not perform well on some detailed textures, such as the nose and eyes in the first row, the windows and doors in the send row, and the curtain, wall picture, and two lamps in the last row. IDM is more effective in reconstructing them and exhibits excellent fine details.

\begin{table}[]
\centering
\caption{Quantitative comparison (PSNR and SSIM) with several baselines on 16$\times$16  $\rightarrow$ 128$\times$128 face super-resolution. Consistency measures the MSE $\left(\times 10^{-5}\right)$ between LR and downsampled SR images.}
\label{tab:face}
\begin{tabular}{lccc}
\hline
Method     & { PSNR$\uparrow$} & { SSIM$\uparrow$} & {Consistency$\downarrow$} \\ \hline
PULSE \cite{menon2020pulse}     & 16.88                       & 0.44                        & 161.1                              \\
FSRGAN  \cite{chen2018fsrnet}   & 23.01                       & 0.62                        & 33.8                               \\
Regression \cite{saharia2022image} & 23.96                       & 0.69                        & 2.71                               \\
SR3 \cite{saharia2022image}        & 23.04                       & 0.65                        & 2.68                               \\ \hline
IDM      & \textbf{24.01}                       & \textbf{0.71}                        & \textbf{2.14}                               \\ \hline
\end{tabular}
\end{table}

\begin{table}[]
\caption{Quantitative comparison (PSNR and LPIPS) on LSUN \cite{yu2015lsun} with 16$\times$ SR.}
\label{tab:nature}
\setlength{\tabcolsep}{0.8mm}{
\begin{tabular}{lccc}
\hline
Method        & Cats      & Bedrooms      & Towers \\
        \hline
% Face     & 21.83/0.4600 & 26.76/0.2787 & 26.84/0.2681 &       \\
PULSE \cite{menon2020pulse}     & 19.78/0.5241 & 12.97/0.7131 & 13.62/0.7066      \\
% Car      & 16.30/0.6491 & 19.42/0.3006 & 19.74/0.2830 &       \\
ESRGAN+ \cite{wang2018esrgan}  & 19.99/0.3482 & 19.47/0.3291 & 17.86/0.3132     \\
GLEAN \cite{chan2021glean}    & 20.92/0.3215 & 19.44/0.3310 & 18.41/0.2850      \\ 
IDM     & \textbf{21.52}/\textbf{0.3131} & \textbf{20.33}/\textbf{0.3290} &    \textbf{19.44}/\textbf{0.2549}   \\
% car 20.27/0.3079
\hline
% cat 21.52/0.5492
% bedroom 20.33/0.5467/74.90
\end{tabular}}
\end{table}

\begin{table}[]
% \label{tab:div2k}
\caption{Quantitative comparison of 4$\times$ SR on the DIV2K \cite{agustsson2017ntire} validation set. D+F means the training datasets include both DIV2K and Flicker2K \cite{timofte2017ntire}, and D means that IDM is only trained on DIV2K. {\color{red}Red} and {\color{blue}blue} colors indicate the best and the second-best performance among generative models, respectively.
% The bold values indicate the best results among generative models.
}
\setlength{\tabcolsep}{0.8mm}{
\begin{tabular}{clccc}

\hline
\multicolumn{2}{c}{Method}                                             & Datasets   & PSNR$\uparrow$                     & SSIM$\uparrow$                                      \\ \hline
\multicolumn{2}{c}{Bicubic}                                            & D+F  & 26.7                      & 0.77                                          \\ \hline
\multirow{2}{*}{Reg.-based}   & EDSR \cite{lim2017enhanced}           & D+F    & 28.98                     & 0.83                                         \\
                              & LIIF \cite{chen2021learning}            & D+F    & 29.00                     & 0.89                                       \\ \hline\hline
\multirow{2}{*}{GAN-based}    & ESRGAN \cite{wang2018esrgan}           & D+F   & 26.22                     & 0.75                                       \\
                              & RankSRGAN \cite{zhang2019ranksrgan}     & D+F  & 26.55                     & 0.75                                       \\ \hline
\multirow{2}{*}{Flow-based}   & SRFlow \cite{lugmayr2020srflow}        & D+F   & 27.09                     & 0.76                                       \\
                              & HCFlow \cite{liang2021hierarchical}    & D+F   & 27.02                     & 0.76                                         \\ \hline
Flow+GAN                      & HCFlow++ \cite{liang2021hierarchical}  & D+F    & 26.61                     & 0.74                                       \\ \hline
VAE+AR    & LAR-SR \cite{guo2022lar}               & D+F   & 27.03 & {\color{blue}0.77}   \\ \hline
Diffusion & IDM   & D   & {\color{blue}27.10}    & {\color{blue}0.77}         \\ 
Diffusion & IDM   & D+F   & {\color{red}27.59}   & {\color{red}0.78}         \\ \hline
\end{tabular}
\label{tab:div2k}}
\end{table}

\begin{table*}[h]
\centering
\caption{Quantitative comparison (PSNR$/$LPIPS) of continuous SR results on CelebA-HQ \cite{karras2017progressive} when training on $8\times$ LR-HR pairs. Each method is trained on 8$\times$ face SR.
% All methods use one model for all scales with training range (1, 8]. 
$``-"$ indicates the model is completely invalid with the magnification.}
\begin{tabular}{ccc|ccc}
\hline
Method& \multicolumn{2}{c|}{in-distribution} & \multicolumn{3}{c}{out-of-distribution} \\ 
   & 5.3$\times$      & 7$\times$                & 10$\times$        & 10.7$\times$        & 12$\times$                \\\hline
LIIF \cite{chen2021learning}   & \textbf{27.52}/0.1207   & \textbf{25.09}/0.1678         &  22.97/0.2246    &22.39/0.2276      & 21.81/0.2332                   \\
% LTE     &\textbf{27.59}   & 25.02      &  \textbf{24.21}     &  22.79     &22.48     & 21.76                   \\\hline
% GLEAN   & /    &  /    &  23.45    &  /     & /    &      /             \\
SR3 \cite{saharia2022image}   & $-$    &  21.15/0.1680        &  20.26/0.2856     & $-$     &      19.48/0.3947              \\
IDM   & 23.34/\textbf{0.0526}    &  23.55/\textbf{0.0736}         &  \textbf{23.46}/\textbf{0.1171}    & \textbf{23.30}/\textbf{0.1238}      &  \textbf{23.06 }/\textbf{0.1800}              \\ \hline
\end{tabular}
\label{tab:continuous}
\end{table*}

% \vspace{-1mm}
\subsection{Quantitative Comparisons}

\paragraph{Face Super-Resolution.} Following SR3, we evaluate IDM on 100 face images extracted from CelebA-HQ \cite{karras2017progressive} and compute the PSNR, SSIM \cite{wang2004image}, and Consistency metrics. Table \ref{tab:face} shows the PSNR, SSIM, and Consistency results on the 8$\times$ face super-resolution task. GAN-based models are up to par with human perception when the super-resolution magnification is large \cite{chan2021glean}. Nevertheless, their poor Consistency values show that their SR results deviate from the LR images. Compared with the diffusion model-based SR3, IDM obtains better results in all metrics (0.97 dB higher in PSNR, 0.06 higher in SSIM, and 0.53 lower in Consistency).

% the poor result reveals the deficiency of LR information in Consistency. Compared to the GAN-based PULSE, FSRGAN and Diffusion-based SR3  and IDM achieves the best results in all metrics 
\begin{figure*}[t]
\centering
\includegraphics[width=0.95\textwidth]{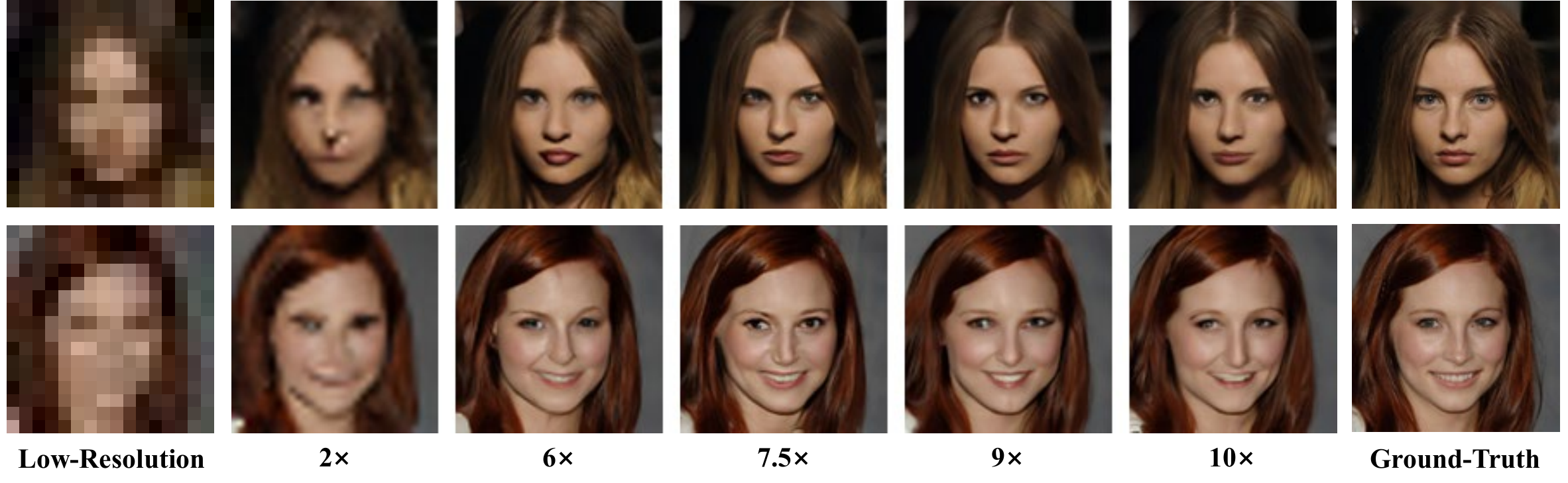} % Reduce the figure size so that it is slightly narrower than the column.
\caption{Visualization of continuous SR results on CelebA-HQ when training on 8$\times$ LR-HR training pairs, where the ground-truth has a resolution of 128$\times$128. We specially select three arbitrary magnifications within the training range (1, 8] and another two out of the range (\emph{i.e.}, 9$\times$ and 10$\times$).}
% As the desired resolution increases, IDM has enriched the smooth details of outputs gradually even if out of the training range (\emph{e.g.} 9 $\times$, 10 $\times$).}
\label{fig:continuous}
\end{figure*}

\begin{figure}[htbp]
\centering
\includegraphics[width=0.5\textwidth]{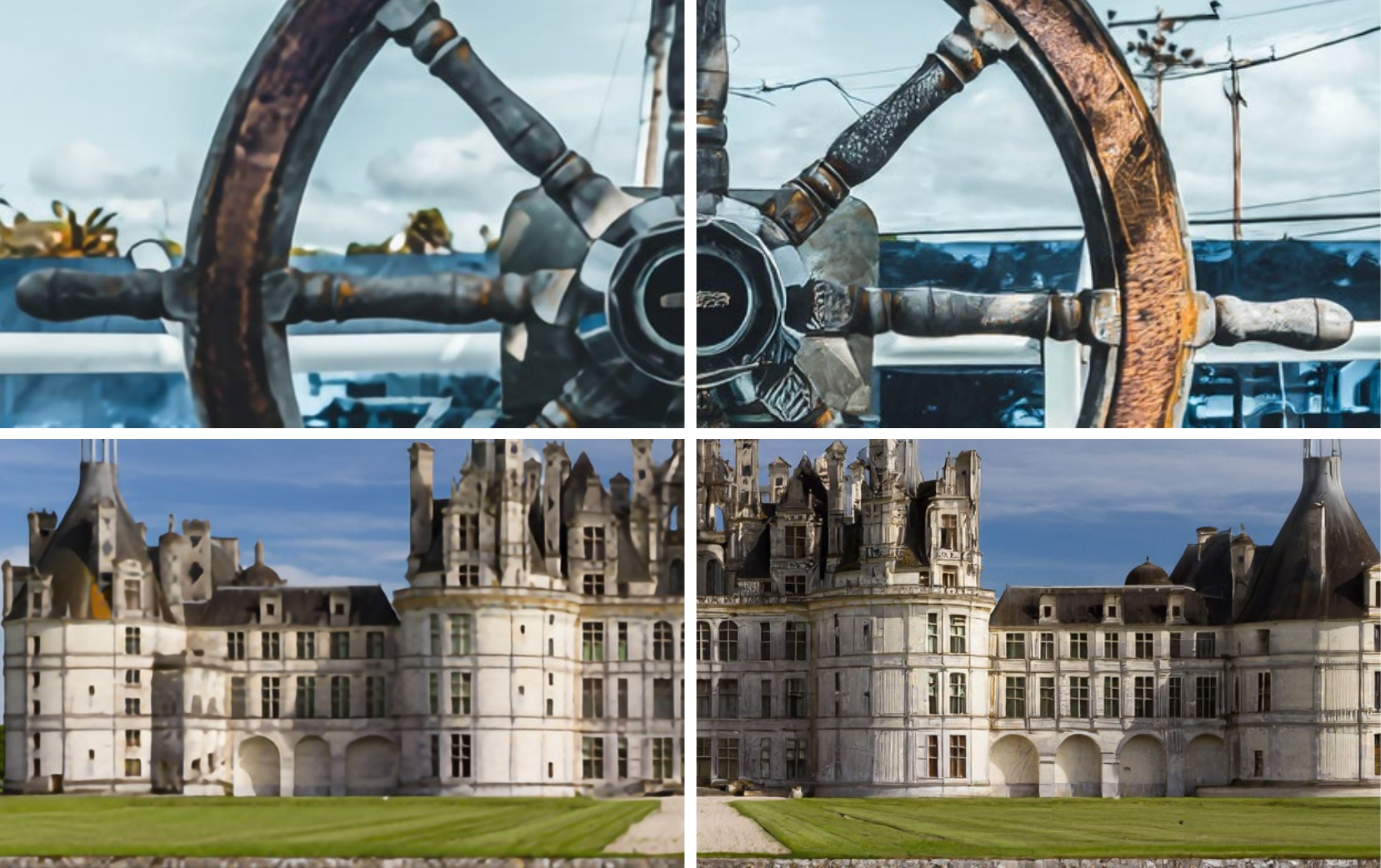} % Reduce the figure size so that it is slightly narrower than the column.
\caption{Two pairs of visual results by the regression-based method LIIF \cite{chen2021learning} (left part of each pair) and our IDM (right part of each pair) for 4$\times$ general scene SR.}
\label{fig:div2k}
\end{figure}

\vspace{-1mm}
\paragraph{Natural Image Super-resolution.} To demonstrate the performance of IDM on natural image SR, we provide the quantitative comparison in Table \ref{tab:nature}. We select 100 images in the validation dataset and compute the average PSNR and LPIPS \cite{zhang2018unreasonable}. Because PULSE generates SR objects incorrectly, its PSNR and LPIPS are significantly lower than other methods. Although GLEAN achieves better results with the pretrained latent banks, IDM outperforms it in all categories, with 0.60 dB, 0.89 dB, and 1.03 dB improvements in PSNR, respectively. Likewise, IDM decreases LPIPS in all categories whose SR results align more with human perception.

% IDM outperforms it in all categories, with 0.60 dB, 0.89 dB, and 1.03 dB improvements in PSNR, respectively. Likewise, IDM decreases LPIPS in all categories whose SR results align more with human perception.
% \begin{table}[]
% \caption{Quantitative comparison (PSNR/LPIPS) on LSUN \cite{yu2015lsun} with 16 $\times$ SR. IDM outperforms other methods in all categories.}
% \label{tab:nature}
% \setlength{\tabcolsep}{0.8mm}{
% \begin{tabular}{lcccc}
% \hline
%             & PULSE        & ESRGAN+ \cite{wang2018esrgan}      & GLEAN \cite{chan2021glean}       & IDM \\ \hline
% % Face     & 21.83/0.4600 & 26.76/0.2787 & 26.84/0.2681 &       \\
% Cat      & 19.78/0.5241 & 19.99/0.3482 & 20.92/0.3215 &    \textbf{21.52}/\textbf{0.3131}   \\
% % Car      & 16.30/0.6491 & 19.42/0.3006 & 19.74/0.2830 &       \\
% Bedroom  & 12.97/0.7131 & 19.47/0.3291 & 19.44/0.3310 &   \textbf{20.33}/\textbf{0.3290}    \\
% Tower    & 13.62/0.7066 & 17.86/0.3132 & 18.41/0.2850 &       \\ \hline
% % cat 21.52/0.5492
% % bedroom 20.33/0.5467/74.90
% \end{tabular}}
% \end{table}

\subsection{Comparison on a General Scene Dataset}
% \paragraph{Qualitative Comparison.} 
We conduct comprehensive comparisons with various prior arts, including regression-based and generative methods, on the general scene dataset DIV2K. EDSR and LIIF are trained with the pixel-wise loss. Fig. \ref{fig:div2k} shows the qualitative comparison with LIIF on the 4$\times$ general scene SR task. LIIF generates clear, high-resolution outputs. However, its simple pixel interpolation leads to an obvious loss of realistic textures.
In Table \ref{tab:div2k}, the results of the GAN-based models (ESRGAN \cite{wang2018esrgan} and RankSRGAN \cite{zhang2019ranksrgan}), flow-based models (SRFlow \cite{lugmayr2020srflow} and HCFlow \cite{liang2021hierarchical}), and mixed generative models (HCFlow++ \cite{liang2021hierarchical} and LAR-SR \cite{guo2022lar}) are from \cite{guo2022lar}.  Table \ref{tab:div2k} demonstrates that our IDM outperforms other generative methods with a significant improvement (0.50dB on PSNR and 0.03 on SSIM). Even with less training data (800 images in ours vs. 2800 images in others), IDM still outperforms prior arts on both metrics.
% achieves the best performance
%  (\emph{e.g.}\lxh{need more professional descriptions})

\begin{figure*}[htbp]
\centering
\includegraphics[width=0.95\textwidth]{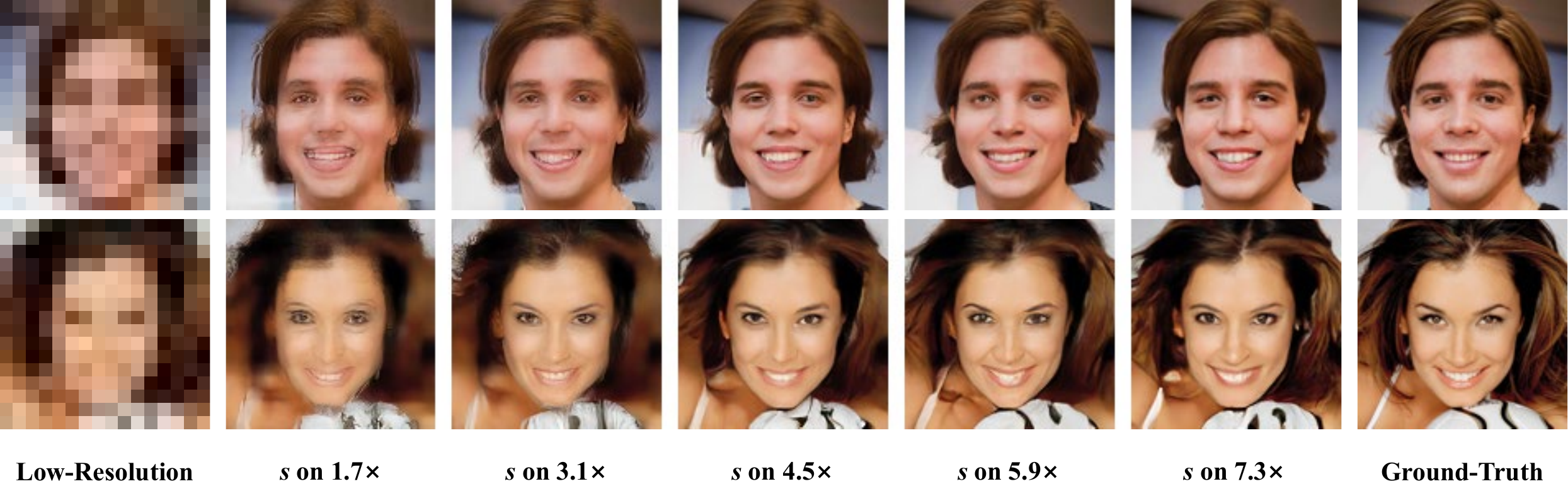} % Reduce the figure size so that it is slightly narrower than the column.
\caption{Visualization with different values of the scaling factor $s$ when training on 8$\times$ face SR, where the ground-truth has a resolution of 128$\times$128, and $s$ takes the values of other magnifications.}
% represents the value of the scaling factor on the corresponding magnification.}
% Four groups of visual results of Regression-based method LIIF \cite{chen2021learning} (The left part of each group) and our IDM (The right part of each group) for 4 $\times$ general image SR.}
\label{fig:scaler}
\end{figure*}

\begin{figure}[htbp]
\centering
\includegraphics[width=0.5\textwidth]{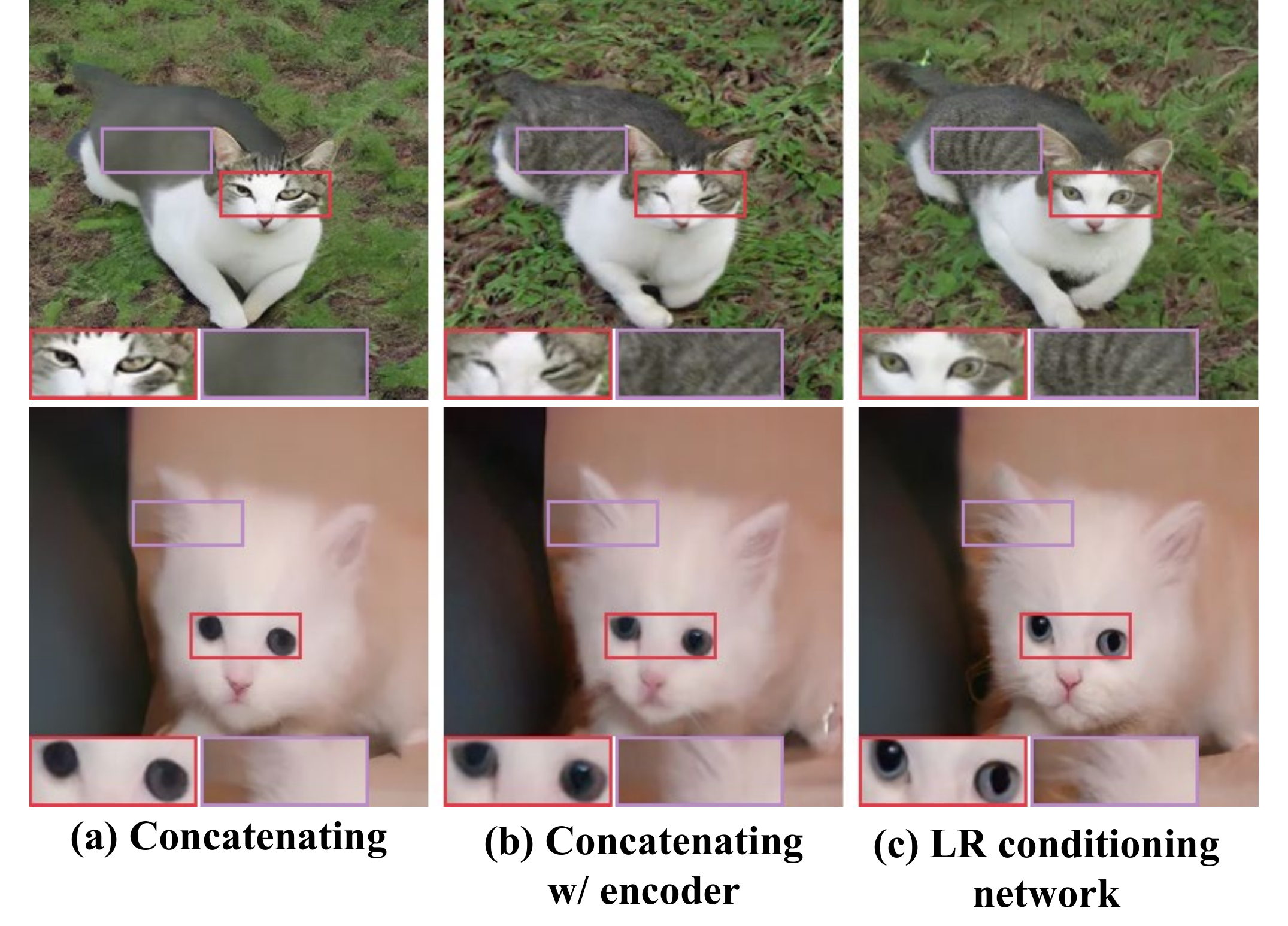} % Reduce the figure size so that it is slightly narrower than the column.
\caption{Effect of the LR conditioning network. (a) Conditioning our model via the concatenating operation in SR3 \cite{saharia2022image}.  (b) Conditioning our model by adding an encoder \cite{li2022srdiff}. (c) Our LR conditioning network.}
\label{fig:ablation}
% \vspace{-3mm}
\end{figure}

% \vspace{-1mm}
\subsection{Comparison of Continuous SR}
\paragraph{Quantitative Results.}
Table \ref{tab:continuous} shows the quantitative comparisons on CelebA-HQ dataset with LIIF and SR3.
% Since GLEAN is not capable for Continuous SR task, we 
% Because of the special network structure, SR3 and GLEAN are not suitable for continuous SR. 
LIIF and IDM are trained within the magnification range (1, 8] and tested on in-distribution and out-of-distribution scales, respectively. For in-distribution scales, although LIIF reports higher PSNR, our IDM exhibits much better performance in terms of LPIPS, demonstrating that the generated images of IDM are much more consistent with human perception. For out-of-distribution scales, IDM outperforms other methods in terms of both PSNR and LPIPS despite the variation of scales. 

% better and more stable results via the variation of scales in terms of both PSNR and LPIPS, which shows the advantage of using implicit neural representation.
\paragraph{Visualization.}
To demonstrate the continuous SR achieved by IDM, we visualize some results with arbitrary testing magnifications when training on 8$\times$ face SR in Fig. \ref{fig:intro_implicit} and Fig. \ref{fig:continuous}. Fig. \ref{fig:intro_implicit} shows that the regression-based models can achieve resolution-continuous results via implicit neural representation, but they suffer from the typical over-smoothing issue (second row).
The generative model (SR3) performs well on the 8$\times$ magnification, consistent with that in training, but it encounters extreme distortions once the magnification changes (third row). 
In contrast, as shown in the fourth row of Fig. \ref{fig:intro_implicit}, IDM successfully synthesizes realistic results with the continuous resolution. In Fig. \ref{fig:continuous}, even if the magnification is out of the training range (1, 8], \emph{i.e.}, 9$\times$ and 10$\times$, IDM still demonstrates outstanding effectiveness in representing continuous SR images.
% We  the resolution-continuous results for 8 $\times$ face SR tasks in Fig. \ref{fig:continuous}. By adjusting the scaling factor, we keep

\subsection{Ablation Studies}
\label{sec:ablation}

\paragraph{Importance of the Scaling Factor.} To demonstrate the significance of the scaling factor $s$, we provide qualitative visual results with different values of $s$ when training on 8$\times$ face SR in Fig. \ref{fig:scaler}. Specifically, we assign $s$ with the value from other specific magnifications. For example, the third column in Fig. \ref{fig:scaler} is obtained using $s$ on 3.1$\times$ face SR. Evidently, for 8$\times$ face SR, using the scaling factor assigned by smaller magnification leads to blurred textures. As the corresponding magnification increases, IDM synthesizes more fine details.  It illustrates that the scaling factor is inclined to allocate more weights to generated features on large-magnification SR. Overall, the scaling factor effectively dynamically adjusts the proportion between LR condition and generated details.
% where the scale factor is assigned by its value in specific magnification. For example, we 
\vspace{-1mm}
\paragraph{Effect of the LR Conditioning Network.} 
We conduct qualitative experiments on Cats with $16\times$ SR to validate the effect of our LR conditioning network. Specifically, we construct two comparison models by replacing the scale-adaptive conditioning network in IDM with two types of conditioning mechanisms that concatenate (1) the upsampled LR image or (2) the LR features encoded by the EDSR encoder with the ground-truth, and feed them to the denoising model, where (1) is adopted by SR3 \cite{saharia2022image}.
% (2)  mechanism introducing an EDSR encoder first to extract LR features as LR condition. 
As shown in Fig. \ref{fig:ablation}(a), directly using the upsampled LR image as the condition often leads to blurred textures. While introducing an encoder to extract features in advance can slightly  alleviate this issue (Fig. \ref{fig:ablation}(b)), it still performs poorly in generating high-fidelity details, such as eyes and hair. In contrast, the proposed scale-adaptive conditioning network develops a parallel architecture providing multi-resolution LR features for the denoising model, to enrich the texture information. Fig. \ref{fig:ablation}(c) shows the superior performance over the others.

% Fig. \ref{fig:ablation} demonstrates how the conditioning mechanism assists IDM in restoring SR outputs with consistent LR features. We compare our scale-adaptive conditioning mechanism with the original concatenating operation \cite{saharia2022image} and a single pattern in the EDSR encoder inputs the LR feature directly. An EDSR encoder can improve texture information significantly from Fig. \ref{fig:ablation}(a) and Fig. \ref{fig:ablation}(b). Although alleviating the over-smoothing problem, this conditioning mechanism still performs poorly in the details' fidelity. As shown in Fig. \ref{fig:ablation}, after adding our scale-adaptive conditioning mechanism with a multi-resolution encoder, IDM enriches the texture information (\emph{e.g.} hair) and makes generated details more realistic (\emph{e.g.} eyes).

\section{Conclusion}
This paper presents an Implicit Diffusion Model (IDM) for achieving high-fidelity image super-resolution with continuous resolution. Specifically, we introduce the implicit image function in the decoding part of the diffusion denoising model. This practical end-to-end framework adopts an iterative process of diffusion denoising and implicit neural representation. We further design a scale-adaptive conditioning mechanism, which takes a low-resolution image as a condition to adjust the proportion between LR information and generated details dynamically.
Extensive experiments illustrate that our IDM exhibits state-of-the-art performance.
% Image Super-Resolution,  one of the hottest topics in computer vision,  has achieved a promising performance due to the advancing properties of generative models. However, they suffer from critical over-smoothing, unnatural artifacts, and fixed magnification, and most of them still rely on additional priors or complicated two-stage pipelines.  To  exploit the merits of both methods in a unified framework,
% A scaling factor is used for We introduce an implicit image function in the upsampling part of the U-Net architecture, leading to a hierarchical architecture of an alternating denoising model.
% The LR conditioning network efficiently encodes LR images to multi-scale features for the iterative denoising steps. And the scaling factor works through Multi-Layer Perceptrons (MLP) to adaptively adjust how much the encoded LR and generated features are expressed.
% There are two types of methods,  regression-based and generative models, each of which has its own advantages and disadvantages. Regression-based methods are intuitive  but  often fail  to achieve high fidelity. 

%%%%%%%%% REFERENCES
{\small
\bibliographystyle{ieee_fullname}
\bibliography{egbib}
}

\end{document}